*Research Article*

# Effects of Data Enrichment with Image Transformations on the Performance of Deep Networks

**Hakan Temiz[1]**

[1]Artvin Coruh University, ORCID: 0000-0002-1351-7565, htemiz@artvin.edu.tr

**Abstract**

*Images cannot always be expected to come in a certain standard format and orientation. Deep networks need to be trained to take into account unexpected variations in orientation or format. For this purpose, training data should be enriched to include different conditions. In this study, the effects of data enrichment on the performance of deep networks in the super resolution problem were investigated experimentally. A total of six basic image transformations were used for the enrichment procedures. In the experiments, two deep network models were trained with variants of the ILSVRC2012 dataset enriched by these six image transformation processes. Considering a single image transformation, it has been observed that the data enriched with 180 degree rotation provides the best results. The most unsuccessful result was obtained when the models were trained on the enriched data generated by the flip upside down process. Models scored highest when trained with a mix of all transformations.*

## 1. Introduction

Although the success of the deep networks depends on much more factors, it is basically directly related to the architecture of the networks and the richness and abundance of the data that they are trained on. The richness and redundancy of the data allows for the deep networks to more balanced learning. It causes robust behavior against the diversity and variances of the input data and undesirable negative conditions such as noise and deformations in the data. In addition, it allows the learning ability of the network to be distributed as homogeneously and evenly as much as possible to all components of the network. In terms of better training the network, it is very important to pass the network through a learning process in which the conditions mentioned above are also included in the process, rather than training the network with a dataset prepared with certain assumptions. For this purpose, a training process should be designed by taking into account the variants of the data and/or various adverse conditions that the network may encounter. In order to provide these conditions, the data should be transformed into a form of a rich training set that considers potential transformations and different situations.

In this study, the effects of data (image) enrichment operations are experimentally examined for the super resolution problem. Six basic image transformation operations are introduced in the experiments for training two distinct deep learning models. The experiments clearly demonstrated the effects (positive or negative) of image transformation operations on the performance of the models.

## 2. Related Studies

SR aims at improving the details while increasing the actual resolution in an image. High resolution images are produced from low resolution images through enlarging them. SR is considered as an inverse problem in which high resolution images would be estimated from low resolution images. The low resolution (LR) images can be produced in simulations with the following formula for a given high resolution (HR) image:

$$x(i,j) = D(B(M(y))) + \eta(i,j) \tag{1}$$

where x and y are respectively low and high resolution images. In the formula, M and B designate warping and blurring operations, respectively, and D indicates downsampling. η denotes the additive noise.

In recent studies there are two main approaches in to super resolve the images: supervised and unsupervised [1] learning. Supervised methods aim at finding a mapping function between (LR) images and (HR) images by introducing the LR and HR image pages in the function. Most algorithms developed in recent years try to find the function that provides the match between low and high resolution images by training them with image patches obtained from single image samples. This kind of training is called as example based learning. Deep networks have shown very superior performance in super resolution problem through this kind of training.

Deep networks outweighed canonical methods and techniques in the super-resolution (SR) problem. Models developed in recent years apply a special methodology based on learning from single image samples. Sophisticated architecture of contemporary deep networks such as Convolutional Neural Networks (CNNs), AutoEncoders (AEs) [2], Generative Adversarial Networks (GANs) is the main reason why these techniques outperform traditional techniques. As per deep convolutional networks, a numerous design features have already been proposed such as residual learning [3][4][5][6], densely connection [7][8], shortcut connection [9][10], iterative dense blocks with double upsampling layers [2], and etc.

Some of CNN that adopted the residual learning approach can be counted as DRCN [11], VDSR [5], DRRN [12]. The EDSR [14] and ESPCN [13] offered the sub-pixel operation in the convolutional layers. FSRCNN [16] implemented a rapid and an efficient processing method. The DenseNet [7] and RDN [8] adopted the idea of densely connected layers. The MemNet [17] proposed the recursive unit and a gate unit in its a special architecture. The LapSRN [18] has used the Laplacian pyramid architecture The EnhanceNet [19] targeted to create more realistic textures. Another special CNNs are the MGEP-SRCNN [21] which introducing a multilabel gene expression programming and the HCNN [20] offering a hierarchical structure.

Very recent studies with GANs have showed an exceptional success in the SR. The SRGAN [22] is proposed to recover better grain details, specifically for large upscaling factors. It has introduced the perceptual loss method composed of an adversarial loss and a content loss. The ESRGAN [23] is derived from the SRGAN architecture. The RTSRGAN [24] has collaborated the power of the ESPCN [13] in real-time processing and the advantages of ESRGAN. DGAN [25] proposed multiple generators in its structure. The GCN [26] offered a collective network design. The SRNTT [30] model has offered a reference based SR to produce more grainer details from reference images by focusing on the information loss on the LR images. The CGAN [27] featured a Laplacian pyramid fashion. A cycle in cycle fashion is recently introduced in the MCinCGAN [28]. The WGAN [29] has offered the Wasserstein GAN architecture. The FG-SRGAN [31] exploited feature-guided SR by stating that it is unfeasible to reconstruct LR images the real reference HR images not exist in the real world. The PGM [33] exploited a probabilistic generative framework offering a low computational cost and noise robustness. The GMGAN [32] integrated the Gradient Magnitude Similarity Deviation (GMSD) metric in its architecture to produce HR images in proportion to the human visual system (HVS). The G-GANISR [34] model introduced the least square loss rather than cross-entropy. Their main aim is to consume all image details without losing any information by progressively increasing the discriminator's charge.

3. Materials and Methods

The model architecture is not only the factor affecting the success of the networks. The training parameters can be counted as another factors that significantly affect their performance. Besides the network architectures, the researchers have studied the training hyper parameters as well to improve the performance. E.g., while some studies have taken into consideration the high learning rate [7][5][6][35], some others preferred to start with relatively small learning rate value [36][18][16]. There are some other approaches for improving the performance, The gradient clipping [5] is just one of these approaches. Majority of other proposals for training hyper parameters are as follows: using different input image size, number of batches, frequency of sample patches from the training image (stride value), etc.

Researchers put lots of effort to excel the success of models by investigating new architectures or optimal set of hyper parameters [37]. Another reason for the success of these techniques is the richness of data. It is well known today that, in general, the more data the more successive models. Apart from these, some of the other factors affecting the performance of the models can be counted as the diversity and redundancy of the training data. In this context, it is desirable to have as many and diverse images for the subject domain in the training set as possible. Some kind of spatial and affine transforms [38] on images can be applied to training datasets for data enrichment.

In example based learning, it is very important to simulate the entire subject domain as comprehensively as possible for deep networks to perform better. However, images may not always be in a standard form. They may appear in any form of spatial or affine transformations (e.g., flat, upside down, rotated, skewed) [38] due to various reasons such as lens distortion of the device or shooting orientation (horizontal, vertical, angular, etc.). Training deep networks also with such transformations will enable networks to be successful in processing images in such conditions. For this purpose, training image dataset should be enriched with various transformations.

In this study, only basic image transformations are considered, although there are many image transformations (translation, projection, or their combinations, etc.). Six different transformations are used for the enrichment: orthogonal rotations of 90, 180 and 270 degrees, and flip operations, left to right, up to down and left to right plus up to down. These operations are depicted in Figure 1.

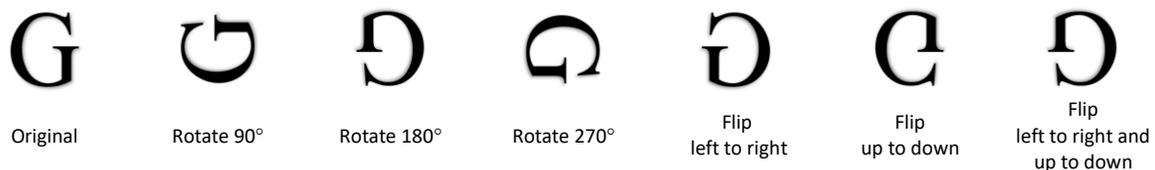

Figure 1. Seven fundamental image transformations used for data enrichment.

### 3.1. Models

In this study, the change in the performances of two deep networks according to the data enrichment in the super resolution problem were experimentally observed. In this context, two deep network structures are exploited in the experiments: DECUSR [2] with three repeating blocks and SRCNN [39]. The both models are of typical CNNs. However, the DECUSR is comprised of twin upsampling layers, and densely connected blocks constructed with the similar idea as in DenseNet [7]. The details of SRCNN and DECUSR models are given in Figure 2 and Figure 3, respectively.

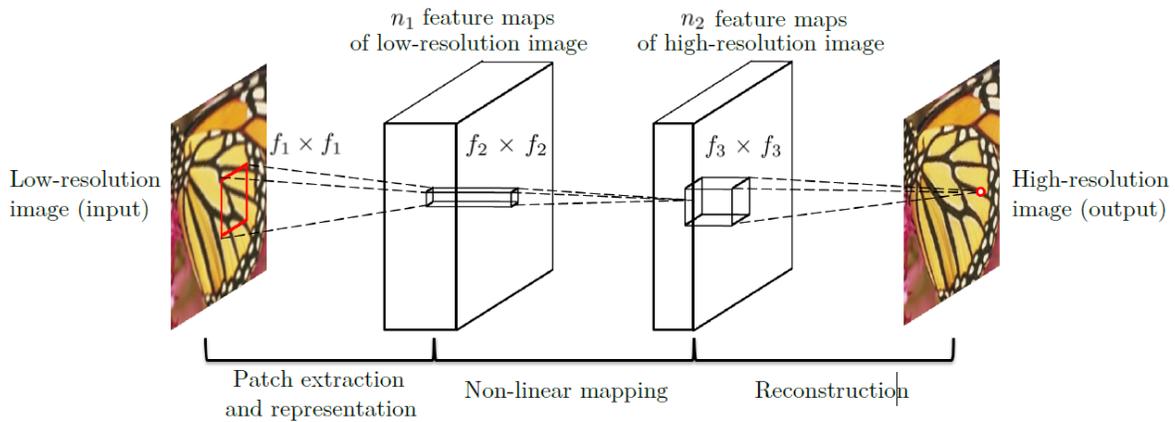

Figure 2. Architecture of SRCNN [39] model.

### 3.2. Training and Dataset

The both models were trained first with the original dataset without any enrichment, and then, trained with the same dataset but this time applying one of the above-mentioned fundamental image transformations. Another training procedure was also performed with the same dataset by applying all transformations together. In total, seven different training procedures were performed.

50 thousand images from the ILSVRC2012 [40] dataset, which comprises of 60 thousand images in total, were introduced in the training procedure for 2 scale factor. The training procedure has carried on for 10 epochs at most. The early stopping procedure is also used to prevent the models from overfitting. The performance of the networks was measured with the PSNR and SSIM metrics on the remaining 9999 images —one image in the test set was excluded since it contains only one pixel information. All experiments are performed with Keras [41] and Tensorflow [42] libraries with NVIDIA GeForce RTX 2080 GPU.

### 4. Results

The results are given in Figure 4. The left and right subfigures show the performances of the DECUSR and SRCNN models, respectively. The red line represents the scores in SSIM, whereas the blue line represents scores in PSNR. While the left vertical axis shows the PSNR, the right vertical axis show the SSIM. The first thing that can easily noticed from the figures is that the performances of both models increase noticeably as soon as data enrichment is applied. Looking at both sub-figures, it can be clearly seen that a higher performance is achieved when the original data is enriched with 90 degree rotation of images than that obtained from other singular transformation operations. Of the single operations, the second best performance was obtained from 270 degrees rotation, according to both metrics. According to the SSIM metric, the SRCNN hit its best performance when dataset is enriched through 180 degrees rotation. According to both metrics, amongst single flip operations, the most successful result was obtained when data enriched with horizontal flip (flip left to right).

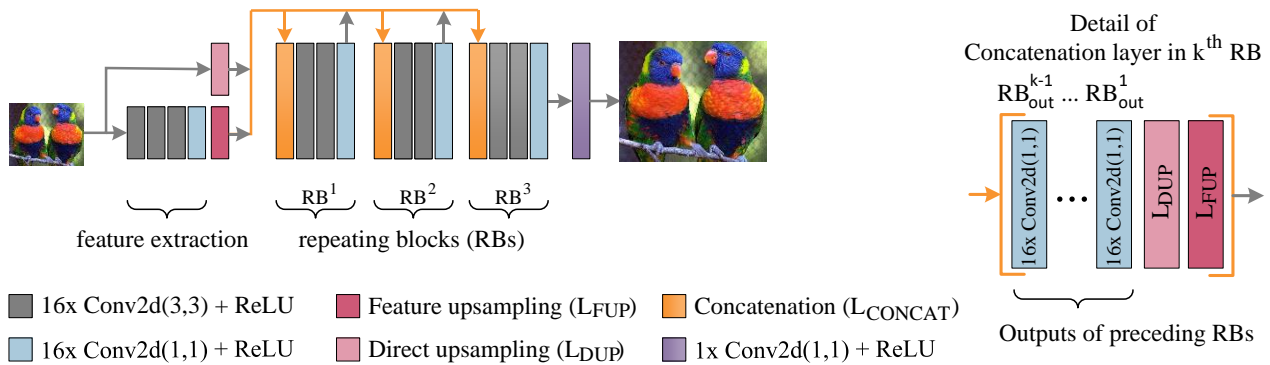

Figure 3. Architecture of DECUSR [2] model with three repeating blocks.

According to the PSNR metric, the worst results of the DECUSR model were obtained when it is trained on enriched data by 180 degrees rotation. The second worst result came from vertical flipping (up to down). Similarly, according to SSIM, the worst result is obtained when it is trained on the data enriched with vertical flip procedure, then, flip on both axis (up to down plus left to right). On the other hand, the worst result for SRCNN according to both PSNR and SSIM metrics was obtained when it is trained on enriched data with vertical flip operation.

As can be seen from the graph, the most successful result for both models was obtained from the training in which all data enrichment processes were included in the process. It is clearly seen how important it is to train deep networks on diversified and enriched data.

In Figure 5, the visual results obtained from both deep learning models from a sample image of the ILSVRC2012 dataset when they are trained on the original and enriched data obtained by applying each of above mentioned enrichment procedures or all together.

The original sample image and the reference piece taken from a certain section of the image are given in the top row. Other rows belong to DECUSR and SRCNN respectively.

They show the visual outcomes of both models when they are trained on each of enriched data. As can be seen from the figure, the visual results obtained are consistent with the quantitative results given in Figure 4. As a result of the training made with the mix of all operations for data enrichment, the models give much sharper, clearer and crisp results.

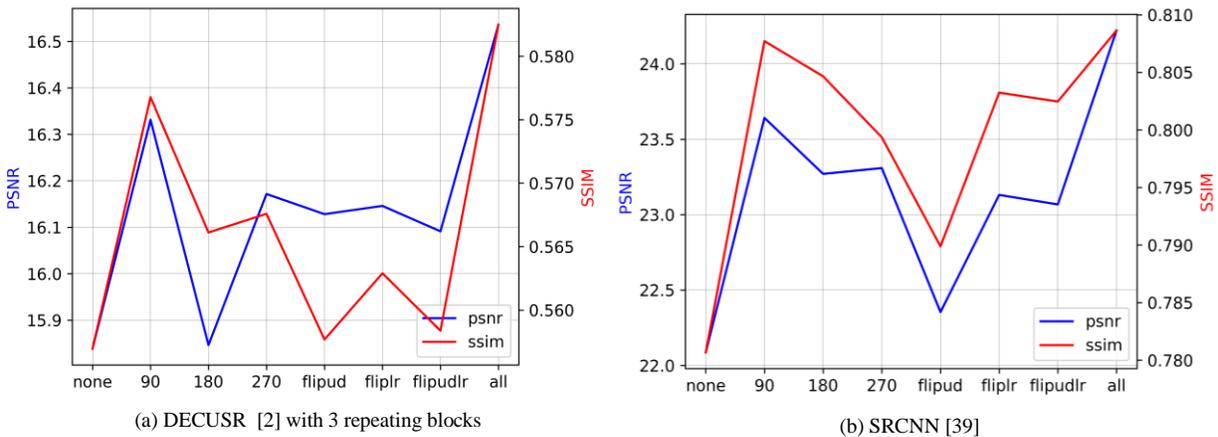

(a) DECUSR [2] with 3 repeating blocks     (b) SRCNN [39]

Figure 4. The performances of models in PSNR and SSIM measures. Subfigure on left is for DECUSR (a) whereas subfigure on right is for SRCNN (b). The abbreviations for flip operations as follows; lr: left to right, ud: up to down, udlr: up to down plus left to right. None indicates training with original dataset wihout any enrichment.

On the other hand, the images obtained from training with the original dataset without any enrichment or augmentation have lower quality. It can be easily observed that many details are lost in the images. In particular, the blurring effect in high-frequency components such as edges is clearly visible. The degradations in the transitions between high and low frequency regions can easily be noticed as well.

## 5.  Conclusion

In this study, it is investigated how the fundamental data enrichment operations affect the performance of deep networks in the super resolution problem. A total of six basic image transformations are exploited for enrichment procedures. For this purpose, DECUSR and SRCNN models were trained with the variants of the ILSVRC2012 dataset each of which were enriched with one of these six image transformation operations and all together. Considering only a single image transformation, it has been observed that the enrichment with 90 degree image rotation provides the best results. The most unsuccessful result was obtained when the models are trained on enriched data generated by flip upside-down operation. The models hit highest scores when they are trained with a mix of all translations.

An important consideration here is to examine the performance achieved by applying a mix of other translations without including those that adversely affect the performance. In the future, it is aimed to observe the outcomes of this concept. Furthermore, other transformations (spatial or affine) will also be included in the experiments.

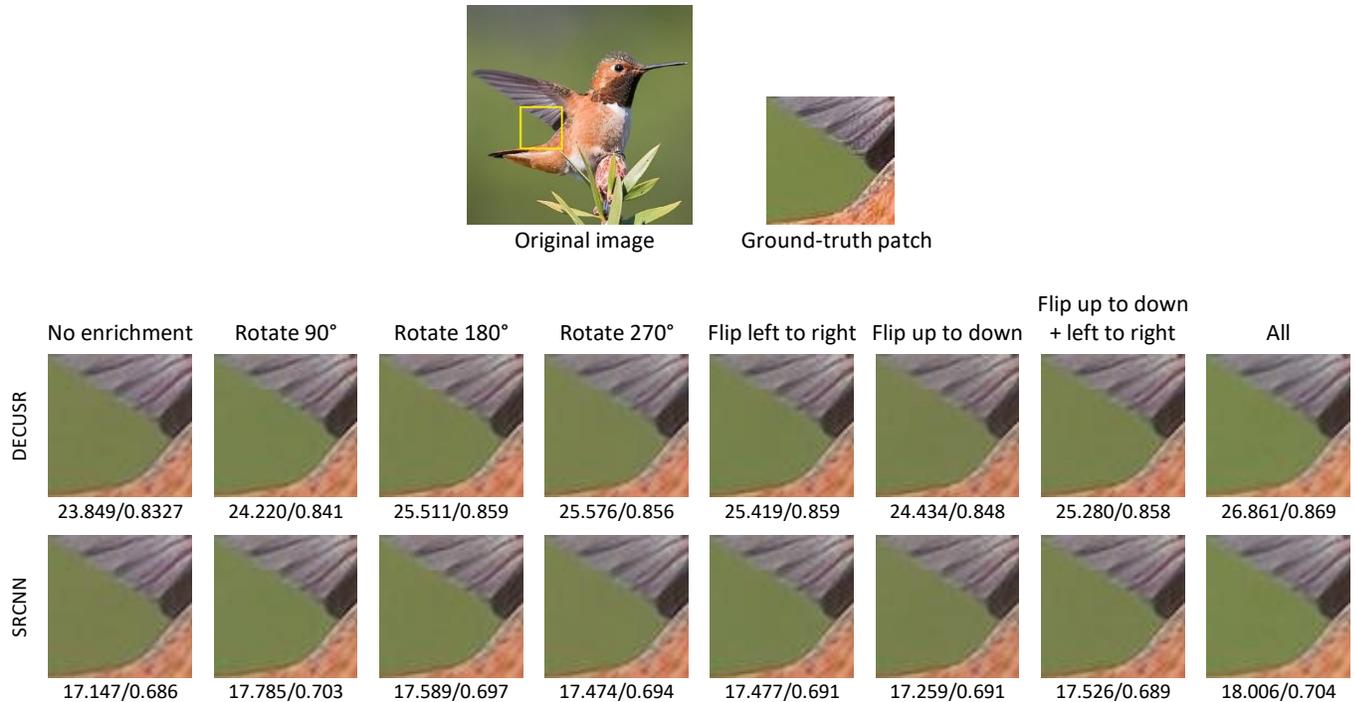

Figure 5. The performances of models in PSNR and SSIM measures. Subfigure on left is for DECUSR (a) whereas subfigure on right is for SRCNN (b). The numeric values shown below the images show the PSNR and SSIM scores.